\documentclass[letterpaper]{article} 
\usepackage{aaai2027}  
\usepackage[hyphens]{url}  
\usepackage{graphicx} 
\usepackage{natbib}  
\usepackage{caption} 
\usepackage{algorithm}
\usepackage{algorithmic}

\usepackage{newfloat}
\usepackage{listings}
\DeclareCaptionStyle{ruled}{labelfont=normalfont,labelsep=colon,strut=off} 
\floatstyle{ruled}
\newfloat{listing}{tb}{lst}{}
\floatname{listing}{Listing}

\usepackage{booktabs}

\usepackage{amsmath}  
\usepackage{amssymb}  
\usepackage{multirow}
\usepackage{placeins}

\title{BRIC-Net: Boundary-Reliable Illumination-Color Interaction \\
	for Remote Sensing Image Deshadowing}

\author{
	Wei Lu,
	Yi Liu,
	Si-Bao Chen\textsuperscript{*}
}
\affiliations{MOE Key Lab of ICSP, IMIS Lab of Anhui, Anhui Provincial Key Lab of Multimodal Cognitive Computation, Zenmorn-AHU AI Joint Lab, School of Computer Science and Technology, Anhui University, Hefei 230601, China\\
	luwei@ahu.edu.cn, e125211073@stu.ahu.edu.cn, sbchen@ahu.edu.cn
}

\begin{document}

\maketitle

\begin{abstract}
Shadows in remote sensing images obscure surface appearance and disrupt radiometric continuity, reducing the reliability of visual interpretation and downstream analysis. Remote sensing image deshadowing is an ill-posed inverse problem that requires spatially varying illumination recovery while preserving chromatic and radiometric consistency in non-shadow regions. Existing methods commonly rely on hard shadow masks for compensation or directly regress RGB intensities. Hard masks may inadequately model gradual penumbra variations and are sensitive to localization errors, often producing residual shadows or halo artifacts; direct RGB regression entangles illumination recovery with chromatic reconstruction and can introduce color casts. To this end, we propose the Boundary-Reliable Illumination-Color Interaction Network (BRIC-Net), which decouples these failures at different representation levels. A Lightness Reliability Prior (LRP) derives reliability-aware guidance from CIELAB statistics. Boundary-Adaptive Gated Mixing (BAGM) performs gated interpolation between shallow RGB and lightness features around uncertain transitions, while Spatial-Channel Mutual Modulation (SCMM) coordinates deeper spatial and channel responses for appearance-preserving illumination recovery. BRIC-Net achieves 29.46~dB full-image peak signal-to-noise ratio (PSNR) on AeroDS-Syn and 27.96~dB on SRGTA. It also obtains the lowest Perception-based Image Quality Evaluator (PIQE) scores on AISD and AeroDS-Real. Region-wise evaluations and component ablations further support its effectiveness in shadow recovery and non-shadow preservation. 
\end{abstract}

\begin{links}
	\link{Code}{https://github.com/AeroVILab-AHU/BRIC-Net}
\end{links}

\section{Introduction}

\begin{figure}[t] \centering
	\includegraphics[width=1\linewidth]{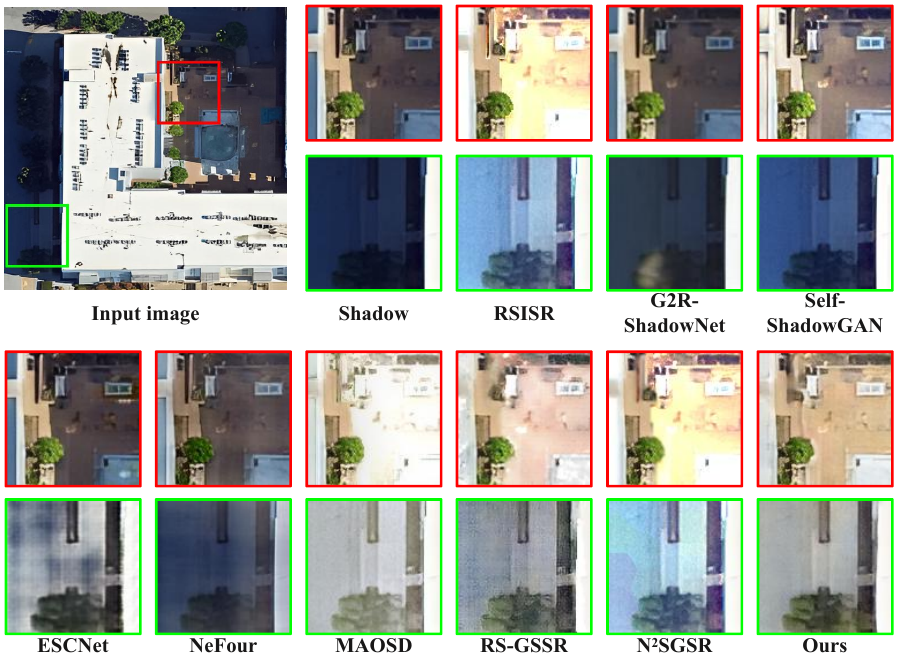}
	\caption{Shadow removal in regions with gradual illumination transitions and heterogeneous surface colors on AeroDS-Real~\cite{Lu2026AeroDeshadow}. Red boxes indicate boundary artifacts, while green boxes highlight local color distortions.}
	\label{fig:fig1}
\end{figure}

Shadows arise when occluders attenuate incident illumination while the underlying surface reflectance remains largely unchanged. In remote sensing images (RSIs), such attenuation can obscure roads, roofs, vehicles, and terrain boundaries over substantial spatial extents. Restoring these regions benefits human interpretation~\cite{Movia2016Shadow} and vision tasks that depend on stable radiometry and structural cues~\cite{Lu2023Robust,Hua2025Multiscale}. RSI deshadowing therefore aims to recover attenuated content while preserving the appearance of the original scene. Because surface reflectance and spatially varying illumination are jointly unknown from a single observation, this task is an ill-posed inverse problem.

This objective is more constrained than generic image brightening. A valid result must satisfy at least four properties: sufficient recovery inside the shadow core, a continuous transition across the shadow boundary, minimal modification of non-shadow regions, and consistent surface color. These requirements are difficult to meet simultaneously in overhead imagery. Large cast shadows may cover several materials, the same image may contain both hard and soft transitions, and sensor-dependent responses produce scene-specific intensity distributions. Consequently, one global gain or one learned RGB transformation can improve average brightness while still producing conspicuous local artifacts.

Traditional methods estimate illumination changes from radiometric models, paired regions, histogram statistics, chromaticity, or gradients~\cite{Silva2018Near,Yu2017New,Murali2013Shadow,Finlayson2006Removal}. Their assumptions are interpretable but often fail under heterogeneous land cover and spatially varying attenuation. Deep networks learn more flexible mappings from paired or unpaired data~\cite{Liu2023Decoupled,Chi2024Neural,Wan2024CRFormer,Chu2025RMMamba}, and recent prior-guided approaches introduce illumination models or decomposition constraints~\cite{Shao2025Generative,Lee2026PhaSR}. Despite this progress, many existing designs still handle transition correction and chromatic preservation through a shared reconstruction mapping~\cite{Wan2024CRFormer,Chu2025RMMamba}, leaving their distinct representation requirements insufficiently modeled.

A major bottleneck in existing pipelines is boundary uncertainty. A binary mask identifies shadow membership but does not explicitly encode the spatially varying attenuation or uncertainty associated with umbra and penumbra~\cite{Shao2025Generative}. Manual annotations and predicted masks additionally contain offsets, omissions, and isolated noise~\cite{Zhu2021Mitigating,Zhang2023Exploiting,Bo2026SARU}. Methods that treat such a mask as deterministic compensation guidance may under-correct transition pixels, leaving residual shadows, or over-correct adjacent non-shadow pixels, producing halos and brightness discontinuities. Figure~\ref{fig:fig1} illustrates these effects. Boundary quality therefore depends not only on the restoration capacity of a network but also on whether its illumination guidance remains reliable near an uncertain mask.

The second challenge concerns illumination-color coupling. RGB channels jointly encode irradiance, material reflectance, and sensor response. Direct RGB regression can thus change inter-channel proportions while increasing intensity, yielding color casts or suppressing local contrast. A cascade that reconstructs brightness before correcting color may propagate first-stage errors~\cite{Wang2024Recreating,Liu2021Shadow}. Conversely, simple feature concatenation in a multi-branch model does not explicitly determine when illumination cues should dominate or when the RGB representation should be retained. Boundary correction and color preservation are therefore coupled at the image level but require different interactions across feature resolution.

These considerations lead to a parallel RGB-lightness formulation. CIELAB lightness provides an intensity-oriented cue with reduced dependence on chromatic channels, whereas the RGB stream retains color and texture. We coordinate these complementary representations according to network depth. At shallow, high-resolution stages, transition locations remain spatially explicit, making gated interpolation suitable for reconciling potentially inconsistent responses. At deeper, lower-resolution stages, enlarged receptive fields provide the contextual and channel dependencies required for appearance-aware illumination recovery. Auxiliary lightness supervision further prevents the illumination stream from becoming an unconstrained latent branch.

We therefore propose the Boundary-Reliable Illumination-Color Interaction Network (BRIC-Net), which realizes this formulation through stage-specific interactions between parallel RGB and lightness representations. Specifically, the Lightness Reliability Prior (LRP) derives distribution-aware lightness guidance by combining histogram-based compensation with boundary calibration. Boundary-Adaptive Gated Mixing (BAGM) performs group-wise interpolation between shallow RGB and lightness features to limit mismatched transfer near uncertain transitions. Spatial-Channel Mutual Modulation (SCMM) uses cross-derived spatial and channel gates in deeper stages to coordinate illumination recovery with RGB appearance. On AeroDS-Syn, BRIC-Net obtains 29.46~dB full-image peak signal-to-noise ratio (PSNR) and 2.47 root mean square error (RMSE); on SRGTA, it reaches 27.96~dB PSNR. It also records the lowest Perception-based Image Quality Evaluator (PIQE) scores on both real-image benchmarks. Collectively, these results show consistent reconstruction gains on paired benchmarks and favorable no-reference perceptual quality on real imagery.

Our contributions are summarized as follows:
\begin{itemize}
	\item We propose BRIC-Net, a stage-specific dual-stream framework for RSI deshadowing. It separates intensity-oriented guidance from RGB appearance reconstruction and coordinates them across network depth, enabling robust joint modeling of reliable illumination recovery alongside comprehensive color preservation.
	\item We instantiate BRIC-Net with complementary, stage-aligned mechanisms. LRP derives distribution-aware lightness guidance, while BAGM performs gated interpolation at shallow, high-resolution stages to reduce sensitivity to uncertain shadow transitions. SCMM mutually modulates deeper spatial and channel responses to coordinate illumination correction with RGB appearance.
	\item Experiments on three RSI benchmarks show that BRIC-Net achieves leading full-image reconstruction accuracy on paired data and the lowest PIQE scores on both real-image benchmarks. Region-wise evaluations and component ablations further support improved shadow-region recovery, non-shadow preservation, and the contribution of the proposed mechanisms.
\end{itemize}

\section{Related Work}

\subsection{Traditional Shadow Removal}

Traditional methods for RSI shadow removal estimate illumination attenuation using physical or statistical assumptions~\cite{Song2014Shadow,Zhang2014Object,Silva2018Near}. Representative approaches infer radiometric relationships between shadowed and sunlit regions~\cite{Chen2007Shadow,Guo2013Paired,Ling2015Shadow}, or employ histogram matching~\cite{Movia2016Shadow}, chromaticity invariance~\cite{Finlayson2006Removal}, and gradient- or boundary-based priors~\cite{Gryka2015Learning}. These methods are interpretable and have limited dependence on training data, but their handcrafted correction assumptions can become unreliable under heterogeneous land cover, inaccurate shadow masks, and spatially varying penumbrae. This limitation motivates using statistical estimates as auxiliary guidance rather than applying them uniformly as final corrections.

\begin{figure*}[t] \centering
	\includegraphics[width=1\linewidth]{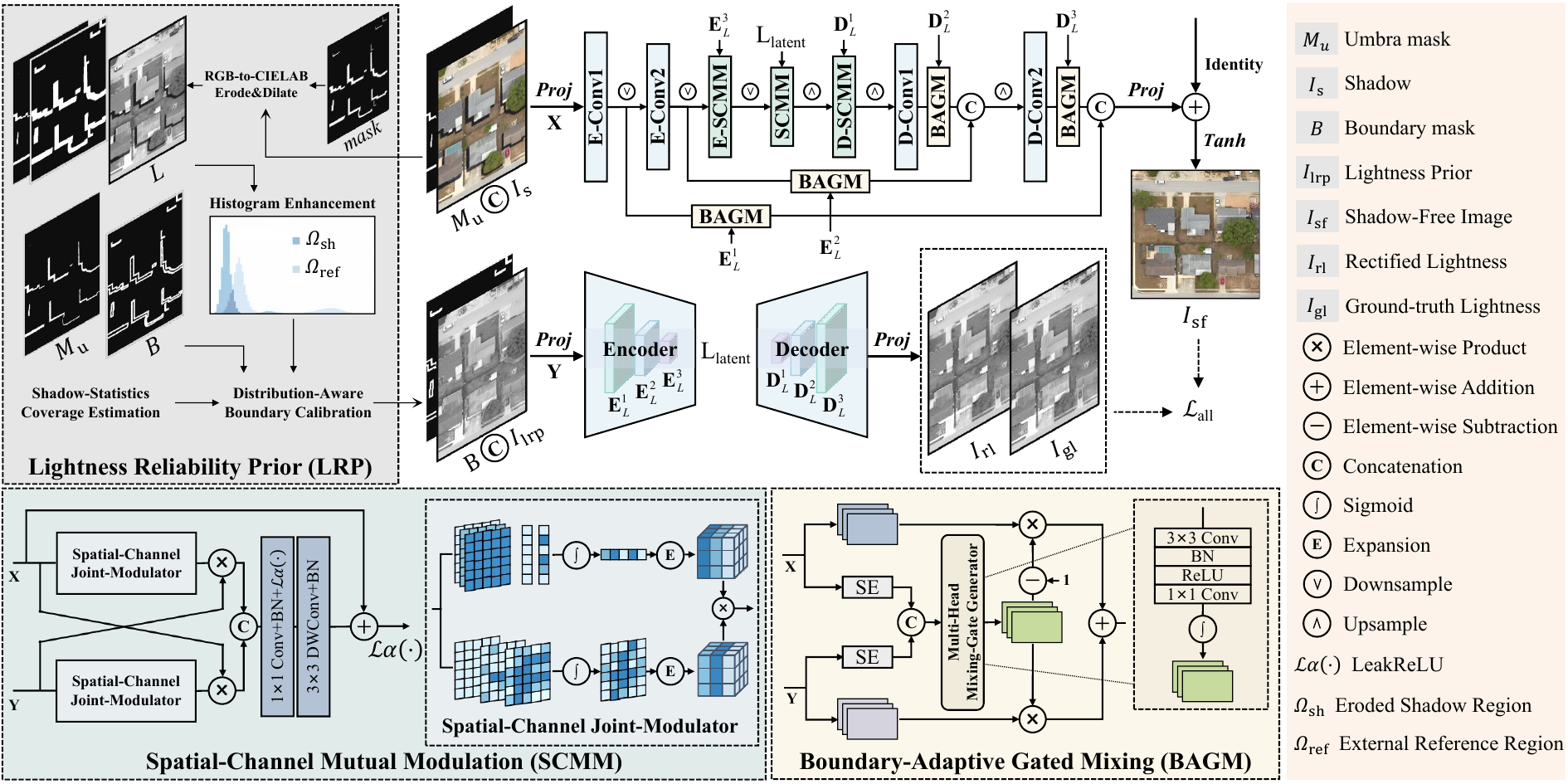}
	\caption{Architecture of BRIC-Net. LRP calibrates uncertain lightness transitions before dual-stream encoding. BAGM mixes shallow RGB and lightness features, while SCMM mutually modulates deeper spatial and channel responses. The RGB decoder predicts the shadow-free image under auxiliary lightness supervision.}
	\label{fig:BRIC_Net}
\end{figure*}

\subsection{Deep Learning-Based Shadow Removal}

Deep shadow removal methods learn nonlinear restoration mappings from paired or unpaired images. Representative directions include generative adversarial network (GAN)-based approaches~\cite{Hu2019Mask,Zhang2020RIS}, attention and transformer architectures~\cite{Wan2024CRFormer,Guo2023ShadowFormer}, state-space models~\cite{Chu2025RMMamba}, frequency-domain restoration~\cite{Chi2024Neural,Dong2024ShadowRefiner}, diffusion models~\cite{Guo2023ShadowDiffusion,Ji2026DCFI}, and joint shadow-aware restoration frameworks~\cite{Bo2026SARU}. These methods improve contextual reconstruction under complex illumination conditions. Nevertheless, architectures centered on a shared RGB reconstruction stream typically optimize transition correction and chromatic preservation within the same representation~\cite{Wan2024CRFormer,Chu2025RMMamba}. This coupling is particularly challenging in overhead imagery, where a single cast shadow may span multiple surface materials and attenuation profiles.

\subsection{Prior-Guided Shadow Removal}
Prior-guided methods incorporate Retinex or intrinsic-image decomposition~\cite{Land1977Retinex,Le2022Physics,Zheng2025Intrinsic}, exposure and illumination modeling~\cite{Fu2021Auto,Shao2025Generative}, geometry-aware compensation~\cite{Zhang2025Shadow}, and physically aligned priors~\cite{Lee2026PhaSR}. Lightness-guided and staged formulations further separate intensity recovery from RGB reconstruction~\cite{Liu2021Shadow,Wang2024Recreating}. These studies establish the value of illumination-oriented representations, but comparatively less attention has been paid to the reliability of such guidance near uncertain boundaries and to how it should interact with learned features across network depth. BRIC-Net focuses on these two aspects by treating statistical lightness compensation as boundary-calibrated guidance, applying gated RGB-lightness interpolation at shallow, high-resolution stages, and performing spatial-channel mutual modulation under auxiliary lightness supervision.

\section{Method}

As shown in Figure~\ref{fig:BRIC_Net}, BRIC-Net separates intensity-oriented guidance from RGB appearance restoration and coordinates the two representations according to feature resolution. We use the following image-formation approximation to motivate this decomposition:
\begin{equation}
	\begin{aligned}
		I_{\mathrm{s}}(x) \approx R(x) \odot E_{\mathrm{s}}(x), \hspace{3mm}
		I_{\mathrm{gt}}(x) \approx R(x) \odot E_{\mathrm{n}}(x),
	\end{aligned}
\end{equation}
where $R$, $E_{\mathrm{s}}$, and $E_{\mathrm{n}}$ denote surface reflectance, shadowed illumination, and target illumination, respectively, and $\odot$ denotes element-wise multiplication. This separates spatially varying illumination recovery from reflectance-associated channel composition. Accordingly, BRIC-Net models illumination using a supervised lightness stream and preserves appearance through an RGB stream.

Given $I_{\mathrm{s}}$ and its binary shadow mask $M$, LRP derives a calibrated lightness prior $I_{\mathrm{lrp}}$ and a boundary band $B$, from which the umbra mask is defined as $M_{\mathrm{u}}=M\setminus(M\cap B)=M\setminus B$. The initial features are
\begin{equation*}
	F_{\mathrm{r}}^{0}=\phi_{\mathrm{r}}([I_{\mathrm{s}},M_{\mathrm{u}}]),
	\qquad
	F_{\mathrm{l}}^{0}=\phi_{\mathrm{l}}([I_{\mathrm{lrp}},B]),
\end{equation*}
where $\phi_{\mathrm{r}}$ and $\phi_{\mathrm{l}}$ are convolutional projections. BAGM operates at shallow encoder-decoder stages with explicit boundary locations. SCMM is applied at deeper stages where contextual and channel dependencies become more informative.

Both streams use three encoder--decoder levels. BAGM produces interaction features after the first two downsampling operations and at their symmetric decoder levels; the encoder-side outputs are retained as cross-stream skip features. SCMM is inserted at the deepest encoder level, the bottleneck, and the first decoder level. The lightness decoder receives its own encoder features, whereas the RGB decoder combines the interaction features with the corresponding encoder-side BAGM outputs. Finally, the RGB branch predicts a residual correction and the lightness branch predicts an auxiliary reconstruction:
\begin{equation}
	\begin{aligned}
		I_{\mathrm{sf}}&=\tanh\!\left(\rho_{\mathrm{r}}([D_{\mathrm{r}}^{1},F_{\mathrm{r}}^{0}])+I_{\mathrm{s}}\right),\\
		I_{\mathrm{rl}}&=\tanh\!\left(\rho_{\mathrm{l}}([D_{\mathrm{l}}^{1},F_{\mathrm{l}}^{0}])\right),
	\end{aligned}
\end{equation}
where $D_{\mathrm{r}}^{1}$ and $D_{\mathrm{l}}^{1}$ denote the final decoder features, and $\rho_{\mathrm{r}}$ and $\rho_{\mathrm{l}}$ are the corresponding output projections. The RGB residual path favors identity preservation in already illuminated regions, while the auxiliary output explicitly constrains the learned lightness representation during optimization.

\subsection{Lightness Reliability Prior}

LRP constructs a statistical lightness prior while accounting for uncertainty near hard-mask boundaries. The shadowed input $I_{\mathrm{s}}$ is transformed into the CIELAB color space, and its lightness channel is denoted by $I_{\mathrm{l}}$.

To estimate a stable shadow-to-reference intensity correspondence, we define two statistical sampling regions directly from the original mask $M$. Eroding $M$ excludes potentially uncertain transition pixels from the shadow statistics, whereas selecting pixels outside its dilated support reduces contamination from nearby penumbrae:
\begin{equation}
	\begin{aligned}
		\Omega_{\mathrm{sh}}
		&= \operatorname{Erode}(M;\mathcal{K}_{5}),\\
		\Omega_{\mathrm{ref}}
		&= \Omega\setminus
		\operatorname{Dilate}(M;\mathcal{K}_{10}),
	\end{aligned}
\end{equation}
where $\mathcal{K}_{5}$ and $\mathcal{K}_{10}$ denote $5\times5$ and $10\times10$ structuring elements, respectively, and $\Omega$ is the image domain. The regions $\Omega_{\mathrm{sh}}$ and $\Omega_{\mathrm{ref}}$ are used to estimate the intensity mapping, with $\Omega_{\mathrm{sh}}$ additionally providing the lightness statistics for boundary calibration.

Independently, the boundary band is constructed from the Canny contour~\cite{Canny1986Computational} of $M$:
\begin{equation}
	\begin{aligned}
		E_M &= \operatorname{Canny}(M),\\
		B &= \operatorname{Dilate}(E_M;\mathcal{K}_{6}),\\
		M_{\mathrm{u}} &= M\setminus B,
	\end{aligned}
\end{equation}
where $\mathcal{K}_{6}$ is a $6\times6$ structuring element. Dilating the contour toward both sides of the annotated boundary produces a spatially coherent transition band and reduces isolated responses associated with irregular mask geometry. Accordingly, $B$ identifies uncertain transition pixels, while $M_{\mathrm{u}}$ retains the interior shadow support used for feature initialization and prior application.

To establish an intensity correspondence, the LUT $T(v)$ matches the cumulative lightness distribution of the eroded shadow region to that of the external reference region:
\begin{equation}
	T(v) = F_{\mathrm{ref}}^{-1}\!\left(F_{\mathrm{sh}}(v)\right),
\end{equation}
where $F_{\mathrm{sh}}$ and $F_{\mathrm{ref}}$ denote the empirical cumulative distribution functions of $I_{\mathrm{l}}$ over $\Omega_{\mathrm{sh}}$ and $\Omega_{\mathrm{ref}}$, respectively. 

The regions $\Omega_{\mathrm{sh}}$ and $\Omega_{\mathrm{ref}}$ provide the distributional statistics independently of $B$ and $M_{\mathrm{u}}$. Both regions are used to estimate $T$, while $\Omega_{\mathrm{sh}}$ additionally provides the lightness statistics for boundary calibration. In contrast, $B$ and $M_{\mathrm{u}}$ determine where and how the resulting prior is applied.

To avoid applying this mapping uniformly to uncertain transition pixels, Distribution-Aware Boundary Calibration uses shadow-distribution statistics to regulate the correction of uncertain transition pixels. The mean $\mu_{\mathrm{sh}}$ and standard deviation $\sigma_{\mathrm{sh}}$ are computed over the eroded shadow region $\Omega_{\mathrm{sh}}$. Under a Gaussian approximation of the lightness distribution within $\Omega_{\mathrm{sh}}$, setting $\alpha=1.96$ yields a nominal 95\% coverage interval:
\begin{equation}
		L_{\text{low}} = \mu_{\text{sh}} - \alpha\sigma_{\text{sh}},  \hspace{5mm} L_{\text{up}} = \mu_{\text{sh}} + \alpha\sigma_{\text{sh}}.
\end{equation}
The full calibrated lightness prior $I_{\mathrm{lrp}}$ is then constructed piecewise across the image domain:
\begin{equation}
	I_{\mathrm{lrp}}(x)=
	\begin{cases}
		T(I_{\mathrm{l}}(x)), & x\in M_{\mathrm{u}},\\
		Q(I_{\mathrm{l}}(x);x), & x\in B,\\
		I_{\mathrm{l}}(x), & \text{otherwise},
	\end{cases}
\end{equation}
where $Q(v;x)$ selectively calibrates values within the transition band $B$:
\begin{equation}
	Q(v;x)=
	\begin{cases}
		v, & v>L_{\text{up}},\\
		T(v), & L_{\text{low}}\leq v\leq L_{\text{up}},\\
		I_{\mathrm{local}}(x), & v<L_{\text{low}}.
	\end{cases}
\end{equation}
Let $\mathcal{N}_{11}(x)$ denote an $11\times11$ neighborhood centered at $x$, and define
$Z(x)=\sum_{y\in\mathcal{N}_{11}(x)}
\mathbf{1}[I_{\mathrm{l}}(y)\geq L_{\mathrm{low}}]$.
The local fallback is
\begin{equation}
	I_{\mathrm{local}}(x)= 
	\begin{cases}
		\displaystyle
		\frac{
			\sum_{y\in\mathcal{N}_{11}(x)} \mathbf{1}[I_{\mathrm{l}}(y)\geq L_{\text{low}}] I_{\mathrm{l}}(y)
		}{
			Z(x) + \epsilon_{\mathrm{a}}
		}, & Z(x)>0,\\[3mm]
		T(I_{\mathrm{l}}(x)), & Z(x)=0,
	\end{cases}
\end{equation}
where $\mathbf{1}[\cdot]$ is the indicator function and $\epsilon_{\mathrm{a}}$ is a small stabilizing constant. The resulting $I_{\mathrm{lrp}}$ acts as scene-adaptive guidance. Learned projections can retain, attenuate, or refine this prior, so that the handcrafted prior is not directly propagated to the RGB output.

\subsection{Stage-Specific Dual-Stream Interaction}

BRIC-Net uses parallel U-shaped RGB and lightness streams. The RGB stream $F_{\mathrm{r}}$ retains color and texture information, while the lightness stream $F_{\mathrm{l}}$ estimates intensity recovery under auxiliary supervision. BAGM and SCMM exchange information based on feature depth without collapsing the two representations into one unconstrained stream.

\subsubsection{Boundary-Adaptive Gated Mixing}

BAGM operates on shallow, spatially detailed features whose receptive fields still preserve transition locations. Independent squeeze-and-excitation transforms~\cite{Hu2018Squeeze} first recalibrate the two streams:
\begin{equation}
	\hat{F}_m = \mathcal{F}_{\text{SE}}(F_m), \quad m \in \{\text{r}, \text{l}\}.
\end{equation}
A multi-head gate $\Phi$ then predicts $K$ mixing maps from their concatenation. We use $K=8$ heads, with the channel dimension $C$ divisible by $K$. $\Phi$ employs a bottleneck projection before applying a sigmoid activation:
\begin{equation}
	G = \{g_k\}_{k=1}^K = \sigma\!\left(\Phi([\hat{F}_{\text{r}}, \hat{F}_{\text{l}}])\right).
\end{equation}
After partitioning the channel dimension $C$ into $K$ groups, each head interpolates between RGB and lightness responses:
\begin{equation}
	\text{out}_k = \hat{F}_{\text{r},k} \otimes g_k + \hat{F}_{\text{l},k} \otimes (1 - g_k),
\end{equation}
where $\otimes$ denotes element-wise multiplication with broadcasting over each channel group. The sigmoid gate makes each output a bounded interpolation between channel-recalibrated RGB and lightness features, limiting mismatched transfer at shallow boundary-sensitive stages.

\subsubsection{Spatial-Channel Mutual Modulation}

SCMM is applied in deeper stages and the bottleneck, where enlarged receptive fields capture contextual dependencies. Each stream independently generates spatial and channel gates, which are subsequently applied to the opposite stream. The spatial gate $S_m$ is produced by depthwise spatial filtering followed by pointwise projection, whereas the channel gate $C_m$ is inferred from GAP:
\begin{equation}
	\begin{cases}
		S_m = \sigma\!\left(\operatorname{MLP}(\operatorname{DWConv}_{3\times3}(F_m))\right) \\
		C_m = \sigma\!\left(\operatorname{MLP}(\operatorname{GAP}(F_m))\right),
	\end{cases}
\end{equation}
where $\operatorname{DWConv}_{3\times3}$ denotes depthwise $3\times3$ convolution and MLP denotes a multilayer perceptron implemented via pointwise projections.

Cross-applied gates implement bidirectional conditioning:
\begin{equation}
		\tilde{F}_{\text{r}} = F_{\text{r}} \otimes (S_{\text{l}} \otimes C_{\text{l}}),  \hspace{5mm}
		\tilde{F}_{\text{l}} = F_{\text{l}} \otimes (S_{\text{r}} \otimes C_{\text{r}}).
\end{equation}
Here, $F_m\in\mathbb{R}^{C\times H\times W}$, $S_m\in\mathbb{R}^{1\times H\times W}$, and $C_m\in\mathbb{R}^{C\times1\times1}$, where the batch dimension is omitted for clarity. The spatial and channel gates broadcast along their singleton channel and spatial dimensions, respectively.

The modulated responses are concatenated and refined by a fusion function $\Psi$, which applies pointwise channel projection followed by depthwise filtering to aggregate channel information and refine local spatial context:
\begin{equation}
	F_{\text{fuse}} = \Psi([\tilde{F}_{\text{r}}, \tilde{F}_{\text{l}}]).
\end{equation}
The fused response is added to the RGB identity path:
\begin{equation}
	F_{\text{final}} = \text{LeakyReLU}(F_{\text{fuse}} + F_{\text{r}}).
\end{equation}
This operation conditions RGB restoration on intensity-oriented evidence from the lightness stream, while the RGB identity connection limits unnecessary perturbations to appearance representations.

\subsection{Optimization Objective}

The network outputs a shadow-free RGB image $I_{\text{sf}}$ and an auxiliary lightness map $I_{\text{rl}}$, supervised by RGB and lightness targets $I_{\text{gt}}$ and $I_{\text{gl}}$. Specifically, $I_{\text{gl}}$ is derived from the CIELAB $L^*$ channel of the ground-truth image. We use standard $\ell_1$ reconstruction and VGG-19 perceptual terms~\cite{Johnson2016Perceptual}, together with the auxiliary lightness and color-ratio constraints:
\begin{equation}
	\begin{aligned}
		\mathcal{L}_{\text{all}} ={}& \lambda_{\text{r}}\|I_{\text{sf}}-I_{\text{gt}}\|_1
		+\lambda_{\text{aux}}\|I_{\text{rl}}-I_{\text{gl}}\|_1 \\
		&+\lambda_{\text{p}}\sum_k w_k\|\phi_k(I_{\text{sf}})-\phi_k(I_{\text{gt}})\|_1
		+\lambda_{\text{c}}\mathcal{L}_{\text{color}},
	\end{aligned}
\end{equation}
where $\phi_k$ and $w_k$ denote a frozen VGG-19 feature map and its scale weight. The loss weights are set to $(\lambda_{\text{r}},\lambda_{\text{aux}},\lambda_{\text{p}},\lambda_{\text{c}})=(80,40,7,200)$. The auxiliary term provides the lightness stream with an explicit objective rather than allowing it to learn an arbitrary latent representation.

To discourage color casts, $\mathcal{L}_{\text{color}}$ constrains normalized RGB proportions. For $I \in \{I_{\text{sf}}, I_{\text{gt}}\}$, let $\bar I=(I+1)/2$ map the normalized RGB values from $[-1,1]$ to $[0,1]$. Then:
\begin{equation}
	\mathcal{L}_{\text{color}} = \frac{1}{3N} \sum_{i=1}^{N} \left\| \frac{\bar I_{\text{sf},i}}{\sum_{c} \bar I_{\text{sf},i}^c + \epsilon} - \frac{\bar I_{\text{gt},i}}{\sum_{c} \bar I_{\text{gt},i}^c + \epsilon} \right\|_1,
\end{equation}
where $i$ indexes spatial pixels, $c \in \{\mathrm{R},\mathrm{G},\mathrm{B}\}$ indexes channels, and $N$ is the total number of pixels.

By normalizing each RGB vector by its channel sum, the ratio term reduces sensitivity to common intensity scaling and emphasizes relative channel proportions. It therefore complements the intensity reconstruction terms by discouraging chromatic deviations. Because of the stabilizing constant $\epsilon$, the invariance is approximate rather than exact.


\section{Experiments}

\begin{table*}[ht]	\centering	\small
	\setlength{\tabcolsep}{2.4pt}
	\renewcommand{\arraystretch}{1.05}
	\begin{tabular}{lcccccccccccccccccc}	\toprule
		\multirow{4}{*}{Method}
		& \multicolumn{9}{c}{AeroDS-Syn}	& \multicolumn{9}{c}{SRGTA} \\
		\cmidrule(lr){2-10}					\cmidrule(lr){11-19}
		& \multicolumn{3}{c}{Shadow}	& \multicolumn{3}{c}{Non-shadow}	& \multicolumn{3}{c}{All}
		& \multicolumn{3}{c}{Shadow}	& \multicolumn{3}{c}{Non-shadow}	& \multicolumn{3}{c}{All} \\
		\cmidrule(lr){2-4}				\cmidrule(lr){5-7}					\cmidrule(lr){8-10}
		\cmidrule(lr){11-13}			\cmidrule(lr){14-16}				\cmidrule(lr){17-19}
		
		& P$\uparrow$ & S$\uparrow$ & R$\downarrow$	& P$\uparrow$ & S$\uparrow$ & R$\downarrow$
		& P$\uparrow$ & S$\uparrow$ & R$\downarrow$	& P$\uparrow$ & S$\uparrow$ & R$\downarrow$
		& P$\uparrow$ & S$\uparrow$ & R$\downarrow$	& P$\uparrow$ & S$\uparrow$ & R$\downarrow$ \\
		\midrule
		
		RSISR 
		& 22.40 & 0.928 & 16.55
		& 33.66 & \underline{0.990} & 1.86
		& 21.59 & 0.911 & 5.80
		& 20.78 & 0.905 & 18.08
		& 30.44 & 0.954 & 2.29
		& 20.18 & 0.848 & 7.37 \\
		
		G2R-ShadowNet
		& 23.99 & 0.948 & 13.44
		& 27.65 & 0.973 & 3.42
		& 21.68 & 0.908 & 5.60
		& 25.02 & 0.943 & 11.33
		& 30.72 & 0.955 & \underline{2.21}
		& 23.68 & 0.891 & 4.97 \\
		
		Self-ShadowGAN 
		& 25.49 & 0.952 & 12.17
		& 28.92 & 0.978 & 2.55
		& 23.58 & 0.921 & 4.66
		& 25.50 & 0.963 & 10.54
		& 29.38 & 0.955 & 2.57
		& 23.64 & 0.911 & 4.72 \\
		
		ESCNet 
		& \underline{30.42} & \underline{0.974} & 7.27
		& \underline{34.12} & 0.977 & 1.63
		& \underline{28.55} & \underline{0.945} & \underline{2.83}
		& 28.31 & 0.977 & 7.56
		& 29.51 & 0.948 & 2.70
		& 25.45 & 0.919 & 3.91 \\
		
		NeFour 
		& 26.71 & 0.960 & 10.30
		& 24.96 & 0.945 & 4.13
		& 22.47 & 0.897 & 5.19
		& 25.01 & 0.963 & 10.90
		& 25.71 & 0.955 & 4.17
		& 22.08 & 0.910 & 5.69 \\
		
		MAOSD 
		& 25.07 & 0.953 & 12.35
		& 29.20 & 0.986 & 2.37
		& 23.37 & 0.929 & 4.65
		& 25.03 & 0.963 & 10.54
		& \underline{30.73} & \underline{0.958} & \underline{2.21}
		& 23.77 & 0.914 & 4.57 \\
		
		RS-GSSR 
		& 30.35 & \underline{0.974} & \underline{7.22}
		& 28.25 & 0.966 & 2.60
		& 25.97 & 0.929 & 3.42
		& \underline{31.22} & \underline{0.985} & \textbf{5.14}
		& 30.29 & 0.944 & 2.34
		& \underline{27.44} & 0.921 & \underline{2.96} \\
		
		N$^2$SGSR 
		& 27.64 & 0.963 & 9.97
		& 32.93 & 0.977 & \underline{1.47}
		& 26.21 & 0.932 & 3.56
		& 24.79 & 0.959 & 11.44
		& 30.29 & \textbf{0.969} & 3.02
		& 23.01 & \underline{0.924} & 5.42 \\
		
		\midrule
		BRIC-Net (Ours)
		& \textbf{30.60} & \textbf{0.976} & \textbf{6.88}
		& \textbf{36.68} & \textbf{0.994} & \textbf{1.11}
		& \textbf{29.46} & \textbf{0.964} & \textbf{2.47}
		& \textbf{31.48} & \textbf{0.986} & \underline{5.15}
		& \textbf{30.96} & 0.956 & \textbf{2.12}
		& \textbf{27.96} & \textbf{0.935} & \textbf{2.78} \\
		\bottomrule
	\end{tabular}
	\caption{Region-wise quantitative comparison on AeroDS-Syn and SRGTA. P, S, and R denote PSNR, SSIM, and RMSE, respectively. Best and second-best results for each metric and region are shown in bold and underlined, respectively.} \label{tab:paired_results}
\end{table*}

\begin{table}[t]	\centering	\small
	\setlength{\tabcolsep}{4.4pt}
	\renewcommand{\arraystretch}{1.05}
	\begin{tabular}{l ccc ccc}
		\toprule
		\multirow{2}{*}{Method}
		& \multicolumn{3}{c}{AISD}
		& \multicolumn{3}{c}{AeroDS-Real} \\
		\cmidrule(lr){2-4}
		\cmidrule(lr){5-7}
		& E$\uparrow$ & B$\downarrow$ & Q$\downarrow$	& E$\uparrow$ & B$\downarrow$ & Q$\downarrow$ \\ \midrule
		
		RSISR 			& 6.49 & 22.06 & 29.52				& 6.92 & 19.11 & 29.86 \\
		G2R-ShadowNet 	& 6.91 & 19.09 & 29.20				& 6.88 & 19.84 & 31.17 \\
		Self-ShadowGAN 	& 7.02 & 22.08 & 27.48				& 7.23 & 18.29 & 30.60 \\
		ESCNet 			& 6.39 & 23.64 & 27.45				& 7.22 & 23.82 & 29.89 \\
		NeFour 			& 5.97 & 18.58 & \underline{24.90}	& 6.73 & 18.93 & 28.89 \\
		MAOSD 			& \underline{7.03} & 19.14 & 26.69	& 7.09 & 18.56 & \underline{28.08} \\
		RS-GSSR 		& 6.98 & 18.78 & 27.61			& \underline{7.26} & \underline{17.60} & 29.60 \\
		N$^2$SGSR 		& 6.48 & \textbf{16.58} & 29.73		& 7.23 & 19.28 & 31.29 \\ 	\midrule
		BRIC-Net (Ours)
		& \textbf{7.04} & \underline{17.49} & \textbf{23.24} & \textbf{7.28} & \textbf{17.19} & \textbf{25.60} \\
		\bottomrule
	\end{tabular}
	\caption{No-reference comparison on AISD and AeroDS-Real. E, B, and Q denote Entropy-S, BRISQUE, and PIQE, respectively. Best and second-best results for each metric on each dataset are shown in bold and underlined, respectively.}
	\label{table_shadow_removal_onAISD} \vspace{-2mm}
\end{table}

\subsection{Experimental Setup}

\begin{figure*}[t] \centering
	\includegraphics[width=1\linewidth]{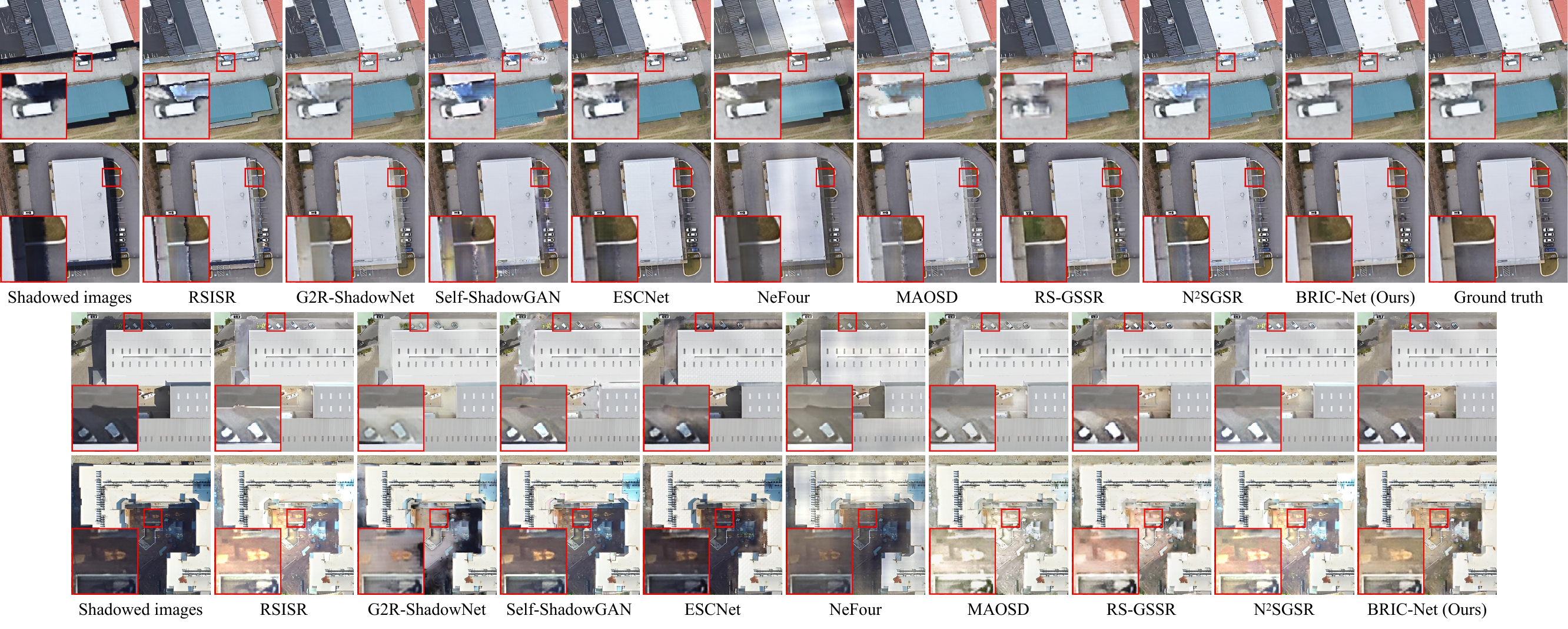}
	\caption{Qualitative comparison on AeroDS-Syn (top two rows) and AeroDS-Real (bottom two rows).}
	\label{fig:BRICNet_AeroDS}
\end{figure*}

\subsubsection{Implementation Details}
BRIC-Net is implemented in PyTorch and trained on a single NVIDIA RTX 3090 Ti GPU. We use the Adam optimizer with a constant learning rate of $1\times10^{-4}$ and a batch size of 4 for 200 epochs. Training images are randomly cropped to $512\times512$ pixels. For paired experiments, BRIC-Net is trained separately on the AeroDS-Syn and SRGTA training splits. AeroDS-Syn, AISD, and AeroDS-Real are evaluated at $512\times512$ pixels, whereas SRGTA predictions and references are evaluated at their original spatial resolutions.

\subsubsection{Datasets}
We evaluate BRIC-Net on three RSI benchmarks. AeroDS~\cite{Lu2026AeroDeshadow} contains 2,000 synthetic training triplets, 260 synthetic test triplets with ground truth, and 260 real test images. SRGTA~\cite{Chu2025RMMamba} contains 1,000 synthetic triplets generated with GTA-V under diverse UAV viewpoints, of which 930 are used for training and 70 for testing. AISD~\cite{Luo2020Deeply} contains 514 shadow image-mask pairs divided into 412 training, 51 validation, and 51 test samples. Only its test split is used for cross-domain evaluation.

\subsubsection{Compared Methods}
We compare BRIC-Net against RSISR~\cite{Silva2018Near}, G2R-ShadowNet~\cite{Liu2021From}, Self-ShadowGAN~\cite{Jiang2023Learning}, ESCNet~\cite{Luo2023Evolutionary}, NeFour~\cite{Chi2024Neural}, MAOSD~\cite{Zhang2025Shadow}, RS-GSSR~\cite{Shao2025Generative}, and N$^2$SGSR~\cite{Bo2026SARU}. All results are reproduced locally. For paired evaluation, each learning-based baseline is retrained on the corresponding benchmark using its official implementation and recommended settings. For cross-domain evaluation, the AeroDS-Syn-trained checkpoint of each learning-based method is directly tested on AISD and AeroDS-Real without target-domain fine-tuning. At inference, dataset-provided masks are supplied only to mask-guided methods, whereas mask-free methods retain their original input configurations.

\subsubsection{Evaluation Metrics}
Following ShadowFormer~\cite{Guo2023ShadowFormer}, we adopt masked-image PSNR and SSIM~\cite{wang2004image} for shadow and non-shadow regions, and report unmasked scores for the full image (ALL). RMSE is computed in the CIELAB space using the corresponding regional masks. For real images, we report shadow-region entropy (Entropy-S) following~\cite{Lu2026AeroDeshadow}, together with full-image BRISQUE~\cite{Mittal2012No} and PIQE~\cite{Venkatanath2015Blind}. All metrics are computed per image and then averaged over each test set.

\begin{figure*}[t]
	\centering
	\includegraphics[width=1\linewidth]{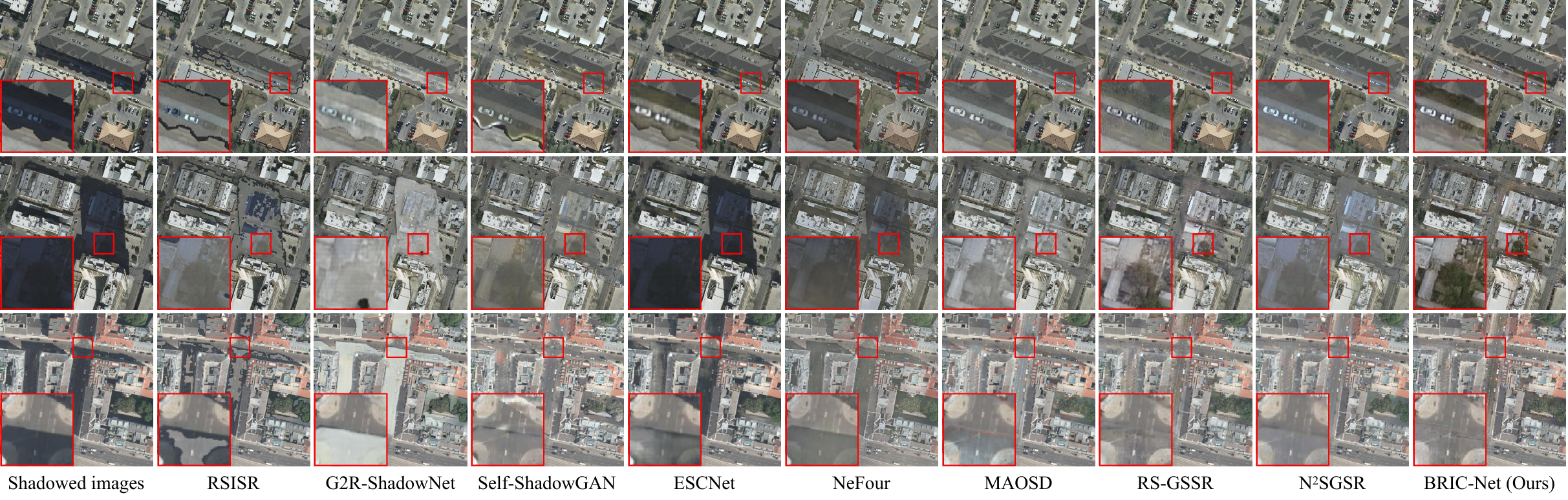}
	\caption{Qualitative comparison on the real-world AISD dataset. Enlarged regions highlight gradual transitions, residual shadows, and local color changes.}
	\label{fig:BRICNet_AISD}
\end{figure*}

\begin{figure}[t]
	\centering
	\includegraphics[width=1\linewidth]{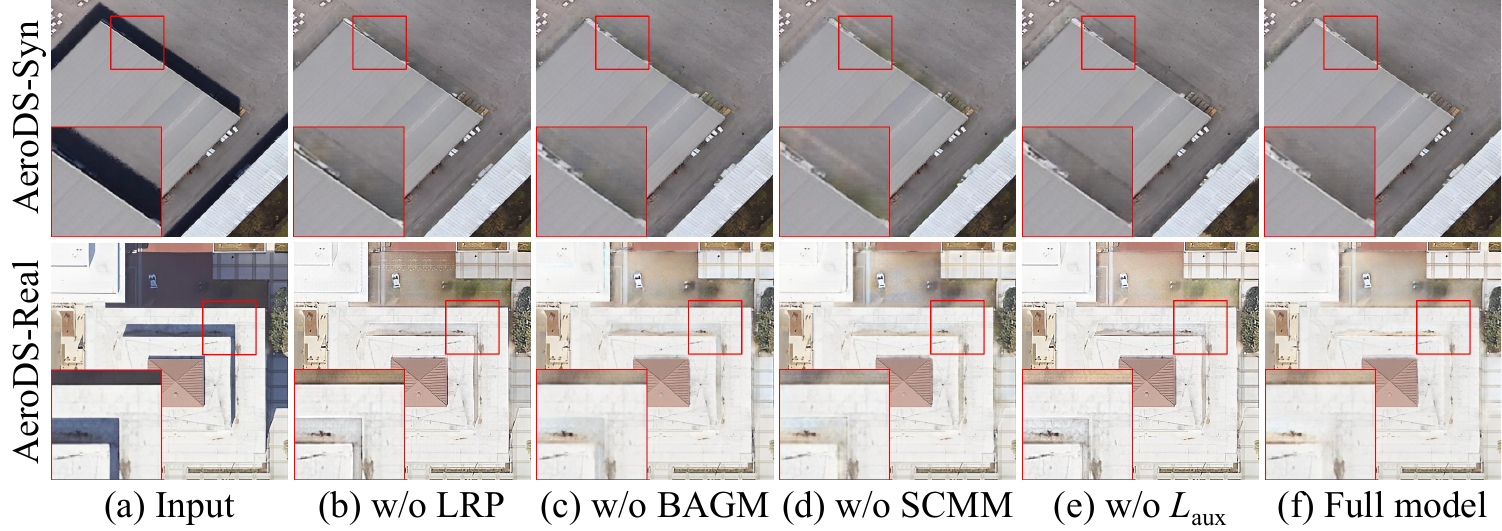}
	\caption{Qualitative ablation on AeroDS-Syn and Real.}
	\label{fig:ablation}
\end{figure}

\subsection{Experimental Results}

\subsubsection{Paired Reconstruction Accuracy}
Table~\ref{tab:paired_results} reports region-wise results on the two paired benchmarks. On AeroDS-Syn, BRIC-Net achieves 29.46~dB PSNR, 0.964 SSIM, and 2.47 RMSE over the full image, improving upon the best competing full-image results by 0.91~dB, 0.019, and 0.36, respectively. Improvements are observed in both shadow and non-shadow regions, indicating that the overall gain is not obtained at the expense of either shadow recovery or illuminated-region preservation.

On SRGTA, BRIC-Net obtains the best full-image PSNR, SSIM, and RMSE. Relative to RS-GSSR, BRIC-Net improves shadow-region PSNR from 31.22 to 31.48~dB, while its RMSE of 5.15 is effectively tied with the best competing value of 5.14. BRIC-Net also achieves the best non-shadow PSNR and RMSE. These results indicate consistent performance across the two spatial regions. On AeroDS-Syn, Figure~\ref{fig:BRICNet_AeroDS} further shows fewer visible transition artifacts while preserving the local appearance of roofs and paved surfaces.

\subsubsection{Real-World Generalization}
Table~\ref{table_shadow_removal_onAISD} reports cross-domain results on unpaired real imagery. On AISD, BRIC-Net obtains the highest Entropy-S (7.04), the lowest PIQE (23.24), and the second-lowest BRISQUE (17.49). On AeroDS-Real, it achieves the best values for all three metrics: 7.28 Entropy-S, 17.19 BRISQUE, and 25.60 PIQE.

Because entropy may also increase with noise or artifacts, it is interpreted jointly with BRISQUE, PIQE, and the visual comparisons in Figures~\ref{fig:BRICNet_AeroDS} and~\ref{fig:BRICNet_AISD}. The results support improved perceptual plausibility under domain shift, but do not establish exact recovery of the unknown radiometry.

\subsection{Ablation Studies}

\begin{table}[t]	\centering	\small
	\setlength{\tabcolsep}{5pt}
	\renewcommand{\arraystretch}{1.08}
	\begin{tabular}{l ccc ccc}
		\toprule
		\multirow{2}{*}{Variant}	& \multicolumn{3}{c}{AeroDS-Syn}	& \multicolumn{3}{c}{AeroDS-Real} \\
		\cmidrule(lr){2-4}	\cmidrule(lr){5-7}
		& P$\uparrow$ & S$\uparrow$ & R$\downarrow$	& E$\uparrow$ & B$\downarrow$ & Q$\downarrow$ \\	\midrule
		
		w/o LRP							& 29.32 & 0.962 & 2.53		& 7.19 & 19.31 & 34.01 \\
		w/o BAGM						& 28.92 & 0.958 & 2.65		& 7.25 & 19.36 & 29.48 \\
		w/o SCMM						& 28.72 & 0.959 & 2.73		& 7.24 & 19.30 & 29.94 \\
		w/o $\mathcal{L}_{\mathrm{aux}}$& 24.96 & 0.931 & 4.05		& 7.25 & 17.69 & 26.80 \\	\midrule
		Full model	& \textbf{29.46} & \textbf{0.964} & \textbf{2.47}	& \textbf{7.28} & \textbf{17.19} & \textbf{25.60} \\  \bottomrule
	\end{tabular}
	\caption{Quantitative ablation on AeroDS-Syn and Real.} \label{table_ablation}
\end{table}

Table~\ref{table_ablation} and Figure~\ref{fig:ablation} analyze the contributions of the proposed components. Removing LRP changes AeroDS-Syn PSNR by only 0.14~dB but increases AeroDS-Real PIQE from 25.60 to 34.01. The corresponding example in Figure~\ref{fig:ablation}(b) retains a dark transition, which is consistent with LRP contributing to uncertain transition calibration.

Removing BAGM lowers synthetic PSNR by 0.54~dB and increases real-image BRISQUE and PIQE by 2.17 and 3.88, respectively. Removing SCMM causes a 0.74~dB PSNR decrease and a 4.34 PIQE increase, while its qualitative example in Figure~\ref{fig:ablation}(d) shows a more visible color change. These observations are consistent with the intended roles of shallow gated interpolation and deep appearance modulation, but they are not direct boundary- or color-specific measurements. Removing $\mathcal{L}_{\text{aux}}$ produces the largest PSNR decrease, indicating that explicit lightness supervision is important to the dual-stream formulation.


\section{Conclusion and Future Work}
We presented BRIC-Net for RSI deshadowing, which maintains separate lightness guidance and RGB appearance streams while coordinating them through stage-specific interactions. Extensive experiments across three benchmarks show that BRIC-Net demonstrates improvements boundary halos and color casts, achieving SOTA paired reconstruction and real-world generalization. Our findings highlight the necessity of stage-specific illumination-color interaction for RSI deshadowing. BRIC-Net currently relies on externally provided shadow masks; future work will investigate joint mask estimation and robustness to mask perturbations.

\newpage
\appendix
\setcounter{page}{1}
\appendix

\twocolumn[
\begin{center}
	\Large \bfseries 
	Supplementary Material for ``BRIC-Net: Boundary-Reliable Illumination-Color Interaction
	for Remote Sensing Image Deshadowing''
	\vspace{3ex} 
\end{center}
]

\section{Implementation Details}
\label{app:implementation}

This section provides the layer-wise implementation details of
BRIC-Net, including module placement, feature dimensions, branch
inputs, output reconstruction, and the complete inference path. These
details specify how the Lightness Reliability Prior (LRP),
Boundary-Adaptive Gated Mixing (BAGM), and Spatial-Channel Mutual
Modulation (SCMM) are integrated into the implemented network.
	
\subsection{Network Organization}

BRIC-Net contains parallel RGB and lightness branches with three
encoder--decoder levels. For a $512\times512$ input, the initial
projections produce 64-channel features. The encoder reduces the
spatial resolution from $512\times512$ to $64\times64$ while increasing
the channel dimension from 64 to 512. The decoder then restores the
feature maps to the input resolution.

The RGB projection receives the four-channel concatenation
$[I_s,M_u]$, where $I_s$ is the shadowed RGB image and $M_u$ is the
umbra mask produced by LRP. The lightness projection receives the
two-channel concatenation $[I_{lrp},B]$, where $I_{lrp}$ is the
calibrated lightness prior and $B$ is the uncertain boundary band.
The RGB branch retains scene color and texture, whereas the lightness
branch provides intensity guidance and identifies uncertain transition
locations.

BAGM is placed at the first two encoder levels and their symmetric
shallow decoder levels, where boundary locations, narrow transitions,
and small structures remain spatially explicit. The encoder-side BAGM
outputs are stored as interaction-aware skip features and reused at the
corresponding decoder scales.

SCMM is placed at the deepest encoder level, the bottleneck, and the
first decoder level. These lower-resolution stages provide a larger
effective receptive field for coordinating spatial context and channel
responses. BAGM and SCMM are instantiated independently at different
scales, without parameter sharing.

The lightness decoder produces an auxiliary restored lightness map,
while the RGB decoder combines encoder features with stored interaction
features to predict an RGB residual. The auxiliary output is supervised
during training, while the RGB reconstruction serves as the
deshadowed result. Table~\ref{table:arch} summarizes the layer-wise
configuration and module placement, and Algorithm~\ref{alg:bricnet}
presents the inference procedure.

\begin{algorithm}[t]
	\caption{BRIC-Net Inference}
	\label{alg:bricnet}
	\small
	\begin{algorithmic}[1]
		\REQUIRE Shadow image $I_s$ and binary shadow mask $M$
		\ENSURE Restored shadow-free image $I_{sf}$
		
		\STATE $(I_{lrp},B,M_u)
		\leftarrow
		\operatorname{LRP}(I_s,M)$
		
		\STATE $F_r^0
		\leftarrow
		\operatorname{Proj}_{RGB}([I_s,M_u])$
		
		\STATE $F_l^0
		\leftarrow
		\operatorname{Proj}_{L}([I_{lrp},B])$
		
		\FOR{$k=1,2$}
		\STATE $(F_r^k,F_l^k)
		\leftarrow
		\operatorname{Encoder}_k
		(F_r^{k-1},F_l^{k-1})$
		
		\STATE $F_{e,mix}^k
		\leftarrow
		\operatorname{BAGM}_{enc,k}(F_r^k,F_l^k)$
		
		\STATE Store $F_{e,mix}^k$
		for the corresponding decoder stage
		\ENDFOR
		
		\STATE $(F_r^3,F_l^3)
		\leftarrow
		\operatorname{SCMM}_{enc}(F_r^2,F_l^2)$
		
		\STATE $(F_{bot}^r,F_{bot}^l)
		\leftarrow
		\operatorname{SCMM}_{bot}(F_r^3,F_l^3)$
		
		\STATE $(D_r^0,D_l^0)
		\leftarrow
		\operatorname{SCMM}_{dec}(F_{bot}^r,F_{bot}^l)$
		
		\FOR{$k=1,2$}
		\STATE $(\widehat D_r^k,D_l^k)
		\leftarrow
		\operatorname{Decoder}_k(D_r^{k-1},D_l^{k-1})$
		
		\STATE $F_{d,mix}^k
		\leftarrow
		\operatorname{BAGM}_{dec,k}
		(\widehat D_r^k,D_l^k)$
		
		\STATE $D_r^k
		\leftarrow
		\operatorname{Fuse}_k
		([F_{d,mix}^k,F_{e,mix}^{3-k}])$
		\ENDFOR

		\STATE $I_{sf}
		\leftarrow
		\tanh\!\left(
		\operatorname{Proj}_{RGB}(D_r^2)+I_s
		\right)$
		
		\STATE $I_{rl}
		\leftarrow
		\tanh\!\left(\operatorname{Proj}_{L}(D_l^2)\right)$
		
		\STATE \textbf{return} $I_{sf}$
	\end{algorithmic}
\end{algorithm}

The RGB residual connection provides a direct route from the input to
the restored output. The auxiliary lightness prediction $I_{rl}$ is
used for training supervision, while $I_{sf}$ is the final reported
deshadowed image.

\begin{table*}[t]
	\centering
	\small
	\setlength{\tabcolsep}{8pt}
	\renewcommand{\arraystretch}{1.08}
	\begin{tabular}{llcccc}
		\toprule
		\textbf{Phase} &
		\textbf{Layer Name (RGB/L)} &
		\textbf{In Ch.} &
		\textbf{Out Ch.} &
		\textbf{Size} &
		\textbf{Key Module} \\
		\midrule
		Input Proj. &
		\texttt{rgb\_proj1} / \texttt{l\_proj1} &
		4 / 2 & 64 & $512^2$ & -- \\
		
		Encoder 1 &
		\texttt{rgb\_enc1} / \texttt{l\_enc1} &
		64 & 128 & $256^2$ & BAGM \\
		
		Encoder 2 &
		\texttt{rgb\_enc2} / \texttt{l\_enc2} &
		128 & 256 & $128^2$ & BAGM \\
		
		Encoder 3 &
		\texttt{scmm\_enc3} / \texttt{l\_enc3} &
		256 & 512 & $64^2$ & SCMM \\
		
		Bottleneck &
		\texttt{bottle\_r} / \texttt{bottle\_l} &
		512 & 512 & $64^2$ & SCMM \\
		
		Decoder 1 &
		\texttt{scmm\_dec1} / \texttt{l\_dec1} &
		512 & 256 & $128^2$ & SCMM \\
		
		Decoder 2 &
		\texttt{rgb\_dec2} / \texttt{l\_dec2} &
		256 & 128 & $256^2$ & BAGM \\
		
		Decoder 3 &
		\texttt{rgb\_dec3} / \texttt{l\_dec3} &
		128 & 64 & $512^2$ & BAGM \\
		
		Output Proj. &
		\texttt{rgb\_proj2} / \texttt{l\_proj2} &
		64 & 3 / 1 & $512^2$ & -- \\
		\bottomrule
	\end{tabular}
	
    \caption{Layer-wise configuration of BRIC-Net for a
	$512\times512$ input. RGB/L denotes the RGB and lightness
	branches. The branch inputs are $[I_s,M_u]$ and $[I_{lrp},B]$,
	respectively. BAGM operates at spatially detailed stages, whereas
	SCMM is used at the deepest encoder, bottleneck, and first decoder
	stages. Modules at different scales use independent parameters.}
	\label{table:arch}
\end{table*}

\subsection{LRP Region Responsibilities and Scale Analysis}
\label{app:lrp_regions}

The regions constructed in LRP serve complementary roles. The eroded
shadow region $\Omega_{sh}$ and the external reference region
$\Omega_{ref}$ provide conservative samples for estimating the
shadow-to-reference lightness correspondence, while the contour-derived
band $B$ supports boundary-specific calibration and
$M_u=M\setminus B$ represents the remaining umbra region. Fixed
$5\times5$ and $10\times10$ structuring elements are used to construct
$\Omega_{sh}$ and $\Omega_{ref}$, respectively.

\begin{figure}[t]
	\centering
	\includegraphics[width=\columnwidth]{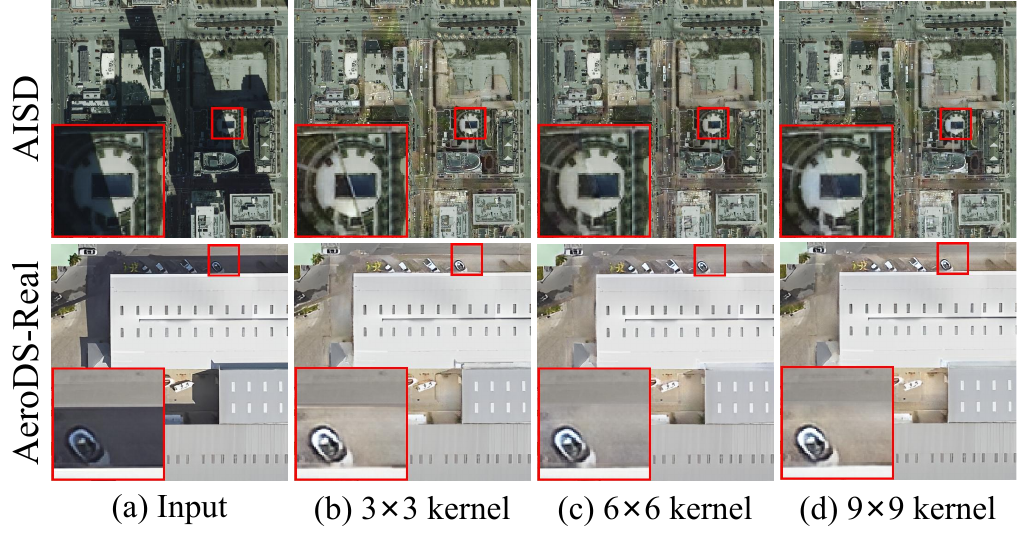}
	\caption{Qualitative sensitivity to the contour-dilation scale used
		to construct the uncertain boundary band $B$. Columns show (a) the
		shadowed input and results obtained using (b) $3\times3$,
		(c) $6\times6$, and (d) $9\times9$ structuring elements. The two
		rows show AISD and AeroDS-Real examples. A narrow band may leave
		residual transitions, whereas a wide band may extend correction
		into neighboring pixels. The $6\times6$ setting provides a more
		balanced result in the displayed regions.}
	\label{fig:boundary_scale}
\end{figure}

We use a $6\times6$ structuring element to dilate the mask contour and
construct $B$. This scale is specifically ablated because it directly
controls the width of the boundary band and consequently affects the
coverage of the shadow transition, the remaining umbra support, and the
boundary-related input supplied to the learned projection. As shown in
Fig.~\ref{fig:boundary_scale}, $\mathcal{K}_3$ tends to leave more
visible transition residuals, whereas $\mathcal{K}_9$ may extend the
correction into neighboring regions. These observations are consistent
with the expected scale trade-off: a narrow band may not fully cover a
gradual transition, while an excessively wide band may include
homogeneous shadow or illuminated content and reduce the retained umbra
support. Table~\ref{table:boundary_scale} shows that
$\mathcal{K}_6$ provides the most consistent overall performance across
the two real-world datasets and the evaluated metrics. We
adopt $\mathcal{K}_6$ in the final configuration.

\begin{table}[t]
	\centering
	\small
	\setlength{\tabcolsep}{5pt}
	\renewcommand{\arraystretch}{1.06}
	\begin{tabular}{c ccc ccc}
		\toprule
		\multirow{2}{*}{Scale} &
		\multicolumn{3}{c}{AISD} &
		\multicolumn{3}{c}{AeroDS-Real} \\
		\cmidrule(lr){2-4}
		\cmidrule(lr){5-7}
		& E$\uparrow$ & B$\downarrow$ & Q$\downarrow$
		& E$\uparrow$ & B$\downarrow$ & Q$\downarrow$ \\
		\midrule
		$\mathcal{K}_3$
		& \textbf{7.06} & 19.08 & 24.99
		& 7.25 & \underline{17.82} & 27.90 \\
		
		$\mathcal{K}_6$
		& \underline{7.04} & \textbf{17.49} & \underline{23.24}
		& \textbf{7.28} & \textbf{17.19} & \textbf{25.60} \\
		
		$\mathcal{K}_9$
		& 6.95 & \underline{17.72} & \textbf{22.66}
		& \underline{7.26} & 17.83 & \underline{26.70} \\
		\bottomrule
	\end{tabular}
	
	\caption{Sensitivity to the contour-dilation scale used to
		construct $B$. E, B, and Q denote shadow-region Entropy-S,
		full-image BRISQUE, and full-image PIQE, respectively. Higher
		Entropy-S and lower BRISQUE and PIQE are preferred. Bold and
		underlined values indicate the best and second-best results. The
		sampling elements $\mathcal{K}_5$ and $\mathcal{K}_{10}$ remain
		fixed in all variants.}
	\label{table:boundary_scale}
\end{table}

\section{Training and Loss Details}
\label{app:training_loss}

Table~\ref{table:training_details} summarizes the optimization
configuration used to train BRIC-Net. Separate models are trained on
SRGTA~\cite{Chu2025RMMamba} and
AeroDS-Syn~\cite{Lu2026AeroDeshadow} because the datasets have different
image characteristics. SRGTA images are randomly cropped into
$512\times512$ patches, whereas AeroDS-Syn samples are already
$512\times512$ and are used at their native resolution. Both models
use Adam with a learning rate of $1\times10^{-4}$, a batch size of 4,
and 200 training epochs.

The complete objective combines RGB reconstruction, lightness
reconstruction, perceptual, and color-ratio constraints. Here, $I_{sf}$
and $I_{gt}$ denote the restored and ground-truth RGB images, while
$I_{rl}$ and $I_{gl}$ denote their corresponding CIELAB lightness maps.
These terms jointly promote luminance recovery and color consistency.

The perceptual term compares the restored image and ground truth through
an ImageNet-pretrained VGG-19 network. Its parameters are frozen during
training, and features from \texttt{relu1\_1} to \texttt{relu5\_1}
measure differences from local edges and textures to higher-level
structures. The selected layers and their fixed scale weights are
reported in Table~\ref{table:loss_vgg}.

\begin{table}[t]
	\centering
	\small
	\setlength{\tabcolsep}{9pt}
	\renewcommand{\arraystretch}{1.05}
	
	\begin{tabular}{lc}
		\toprule
		\textbf{Item} & \textbf{Setting} \\
		\midrule
		Optimizer       & Adam \\
		Learning rate   & $1\times10^{-4}$ \\
		Batch size      & 4 \\
		Training epochs & 200 \\
		Training size   & $512\times512$ \\
		GPU             & NVIDIA RTX 3090 Ti \\
		\bottomrule
	\end{tabular}
	
    \caption{Training configuration of BRIC-Net. Separate models are
	trained on the SRGTA and AeroDS-Syn training splits. SRGTA images
	are randomly cropped into $512\times512$ patches, whereas AeroDS-Syn
	images are used at their native $512\times512$ resolution.}
	
	\label{table:training_details}
\end{table}

\begin{table}[t]
	\centering
	\small
	\setlength{\tabcolsep}{8pt}
	\renewcommand{\arraystretch}{1.05}
	
	\begin{tabular}{lc}
		\toprule
		\textbf{Term / Layer} & \textbf{Weight} \\
		\midrule
		\multicolumn{2}{c}{\textit{Loss weights}} \\
		\midrule
		$\|I_{sf}-I_{gt}\|_1$ & $\lambda_r=80$ \\
		$\|I_{rl}-I_{gl}\|_1$ & $\lambda_{aux}=40$ \\
		$\mathcal{L}_{p}$     & $\lambda_p=7$ \\
		$\mathcal{L}_{color}$ & $\lambda_c=200$ \\
		\midrule
		\multicolumn{2}{c}{\textit{VGG-19 perceptual features}} \\
		\midrule
		S1: \texttt{relu1\_1} & $w_1=0.03125$ \\
		S2: \texttt{relu2\_1} & $w_2=0.0625$ \\
		S3: \texttt{relu3\_1} & $w_3=0.125$ \\
		S4: \texttt{relu4\_1} & $w_4=0.25$ \\
		S5: \texttt{relu5\_1} & $w_5=1.0$ \\
		\bottomrule
	\end{tabular}
	
	\caption{Loss weights of BRIC-Net and the frozen VGG-19 layers with
		their fixed weights used in the perceptual loss $\mathcal{L}_{p}$.}
	\label{table:loss_vgg}
\end{table}

\section{Dataset Examples}
\label{app:dataset_examples}

Representative samples from the four evaluated datasets are presented
below to illustrate their visual characteristics and experimental
roles. SRGTA and AeroDS-Syn provide paired shadowed images, binary
shadow masks, and shadow-free ground truths, supporting supervised
training and full-reference evaluation. AeroDS-Real~\cite{Lu2026AeroDeshadow}
and AISD~\cite{Luo2020Deeply} contain real-world shadow images without
paired shadow-free references and are used to evaluate cross-domain
generalization. These datasets cover diverse aerial scenes,
surface materials, shadow scales, boundary patterns, and illumination
conditions.

\begin{figure}[t]
	\centering
	\includegraphics[width=\columnwidth]{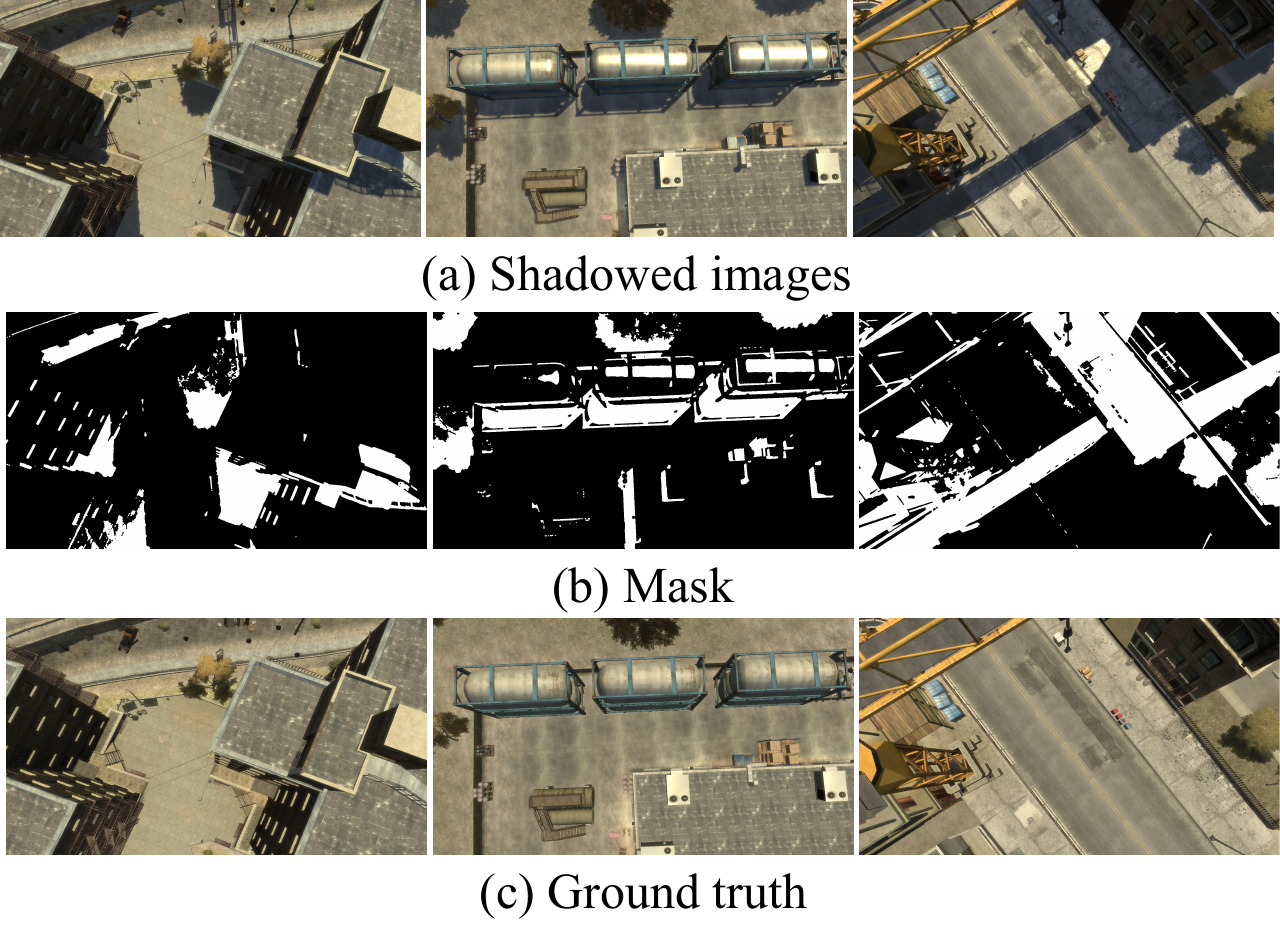}
	\caption{Representative samples from SRGTA dataset. The examples include
		complex shadow geometry, small structures, and shadows extending
		across different surfaces.}
	\label{fig:dataset_srgta}
\end{figure}

\subsection{Paired Synthetic Datasets}

\begin{figure}[t]
	\centering
	\includegraphics[width=\columnwidth]{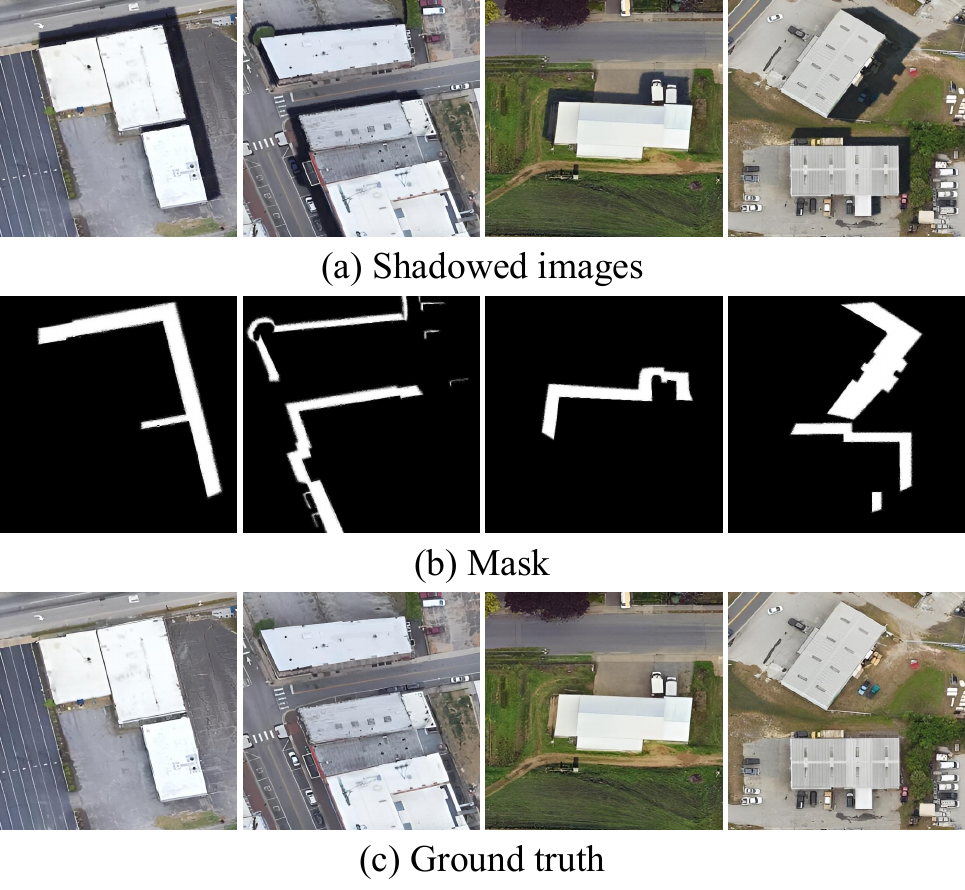}
	\caption{Representative samples from AeroDS-Syn dataset. The examples cover
		different aerial scenes, shadow extents, boundary transitions,
		surface materials, and local structures.}
	\label{fig:dataset_aerods_syn}
\end{figure}

Figures~\ref{fig:dataset_srgta} and
\ref{fig:dataset_aerods_syn} show representative examples from SRGTA
and AeroDS-Syn, respectively. Their paired data enable direct comparison
between shadowed inputs and shadow-free targets. The selected examples
include different scene layouts, shadow extents, boundary profiles,
surface materials, and local structures, thereby reflecting the variety
of conditions considered during training and quantitative evaluation.

\subsection{Real-World Datasets}

\begin{figure}[t]
	\centering
	\includegraphics[width=\columnwidth]{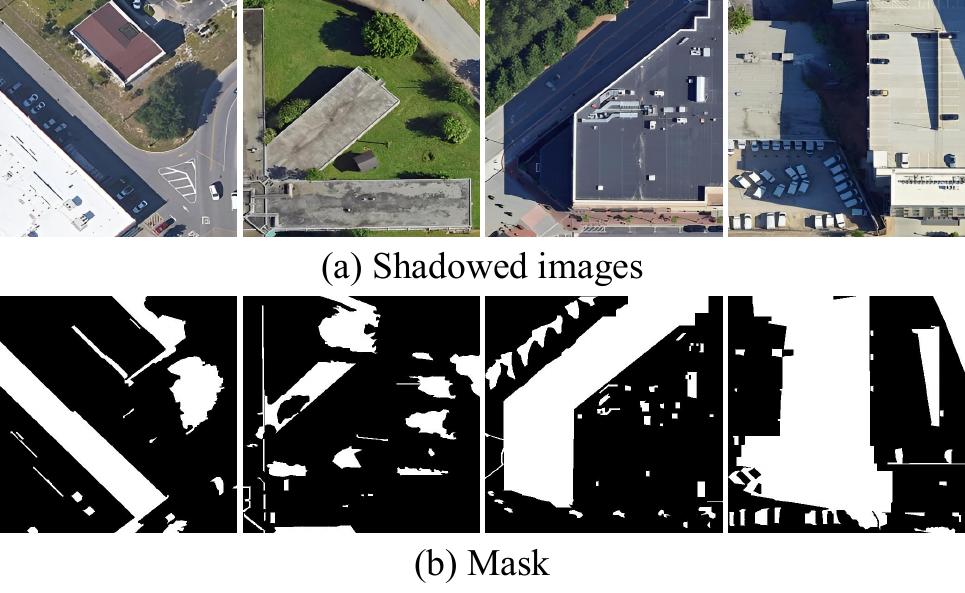}
	\caption{Representative samples from AeroDS-Real dataset. The examples cover
		buildings, roads, vegetation, heterogeneous surfaces, and shadows
		with different spatial extents and boundary profiles.}
	\label{fig:dataset_aerods_real}
\end{figure}

\begin{figure}[t]
	\centering
	\includegraphics[width=\columnwidth]{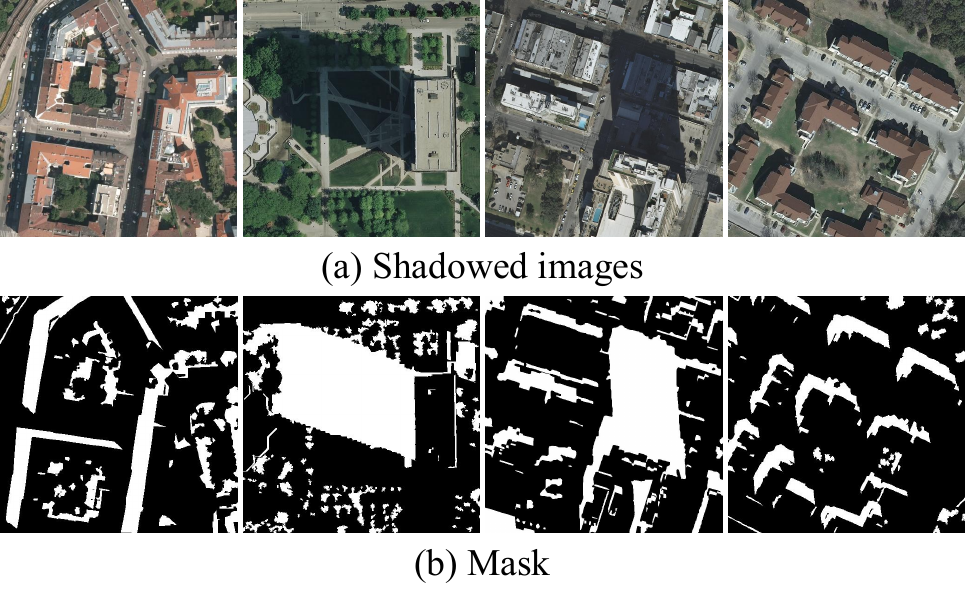}
	\caption{Representative samples from the AISD dataset. The examples
		include dense structures, vegetation-rich regions, large irregular
		shadows, and small or narrow shadows under cross-domain evaluation.}
	\label{fig:dataset_aisd}
\end{figure}

Figures~\ref{fig:dataset_aerods_real} and
\ref{fig:dataset_aisd} present examples from AeroDS-Real and AISD,
respectively. Because paired shadow-free targets are unavailable, the
samples contain only observed shadow images and their binary masks.
Compared with the synthetic datasets, these images exhibit more
complicated acquisition conditions, heterogeneous land cover, and less
regular illumination transitions. They are therefore used to examine
whether the evaluated methods remain effective when applied to real and
previously unseen shadow distributions.

\section{Additional Qualitative Results}
\label{app:qualitative}

\begin{figure*}[t]
	\centering
	\includegraphics[width=\textwidth]{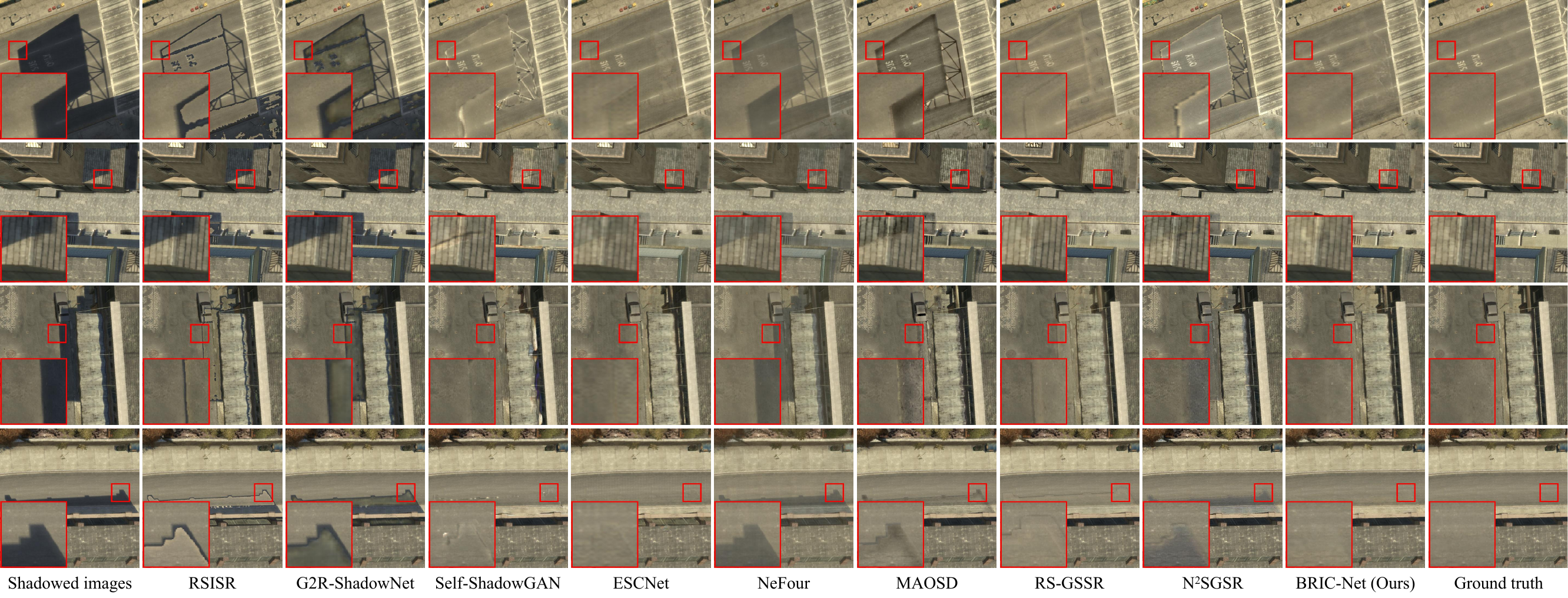}
	\caption{Additional qualitative comparison on the paired SRGTA
		dataset. The final column provides the ground truth. The selected
		scenes contain complex object contours and shadows cast across
		surfaces with different appearances. In the shown examples,
		BRIC-Net more closely follows the target transitions while
		preserving surrounding details, consistent with the region-wise
		full-reference results.}
	\label{fig:gta}
\end{figure*}

\begin{figure*}[t]
	\centering
	\includegraphics[width=\textwidth]{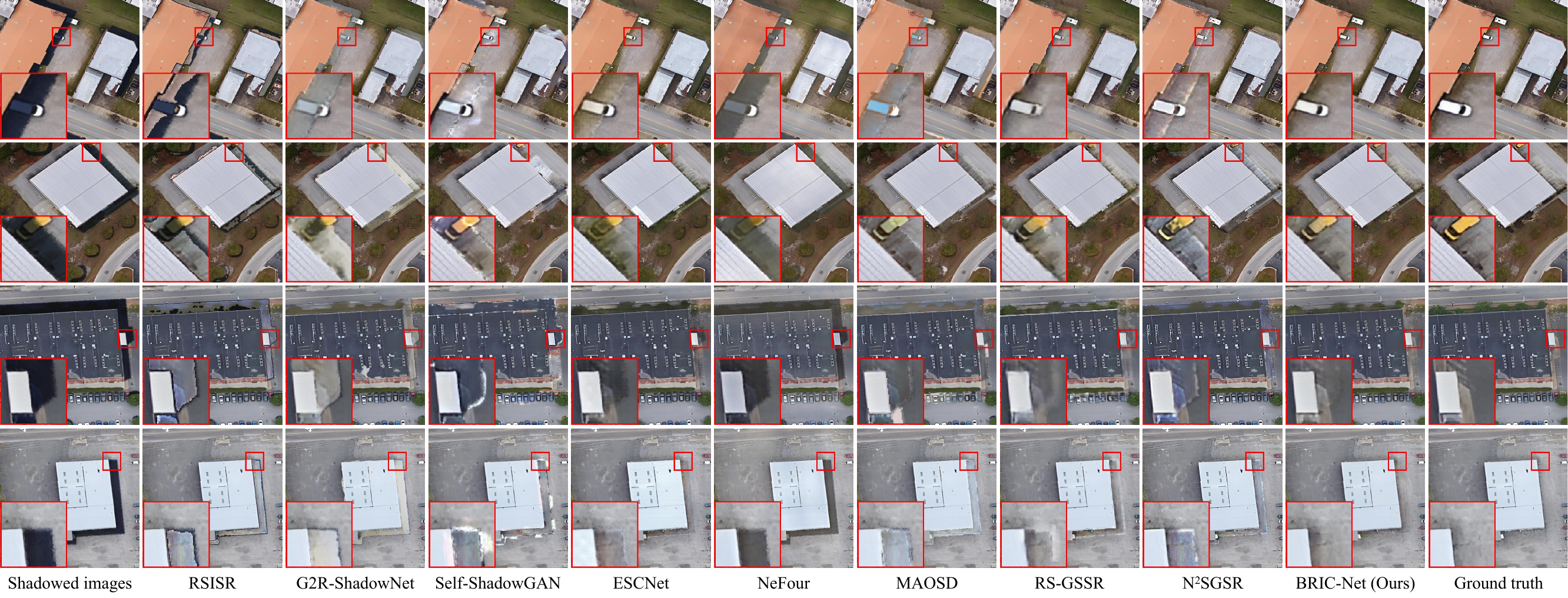}
	\caption{Additional qualitative comparison on the paired AeroDS-Syn
		dataset. The ground-truth column permits direct comparison with the
		target. The displayed shadows cover rooftops, paved surfaces, and vehicles,
		making a uniform intensity correction insufficient. BRIC-Net shows fewer conspicuous transition artifacts
		and closer local color agreement with the target in the presented
		examples, in line with the quantitative paired evaluation.}
	\label{fig:syn}
\end{figure*}

\begin{figure*}[t]
	\centering
	\includegraphics[width=\textwidth]{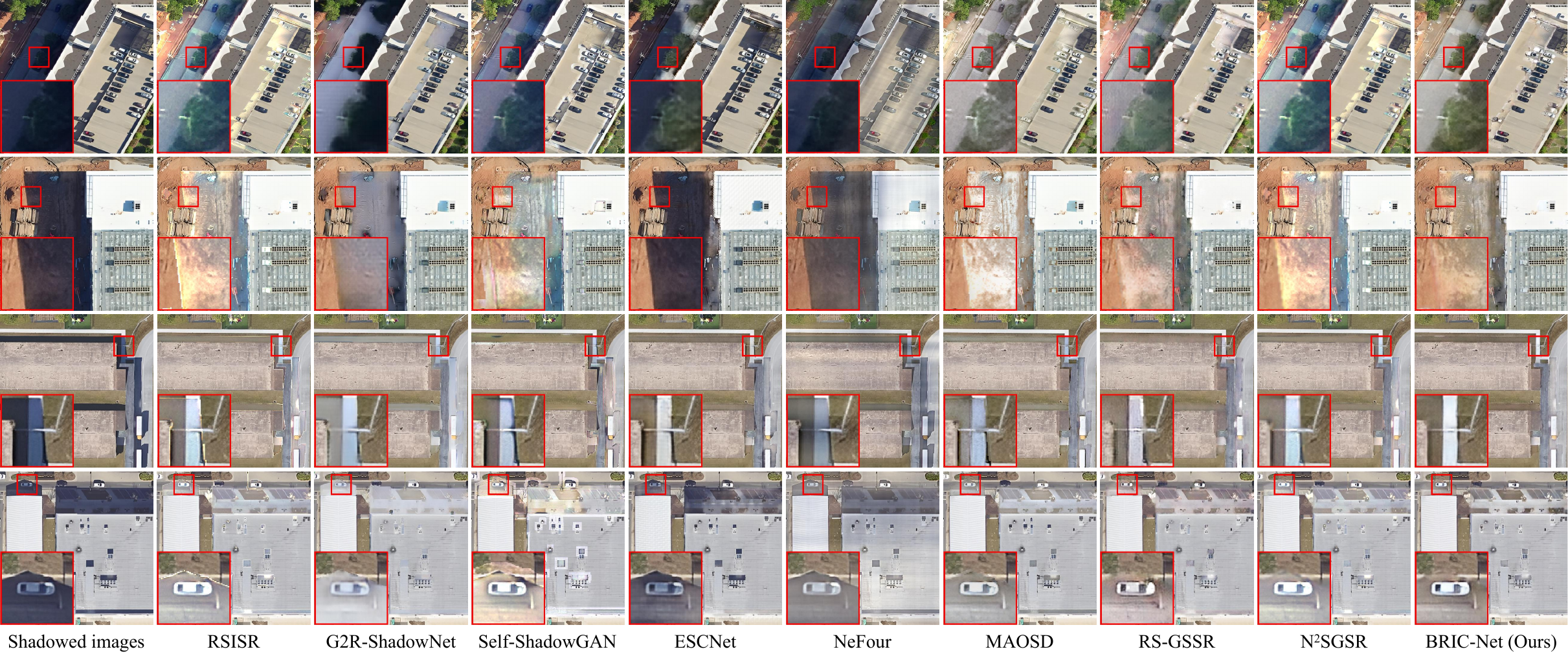}
	\caption{Additional qualitative comparison on the AeroDS-Real
		dataset. From left to right, the columns show the shadowed input,
		competing methods, and BRIC-Net. The enlarged regions emphasize
		real-world shadows with gradual transitions across vegetation,
		soil, roads, roofs, and vehicles. In the displayed cases, BRIC-Net
		produces visually smoother transitions and more consistent local
		appearance while retaining identifiable scene structures.}
	\label{fig:real}
\end{figure*}

\begin{figure*}[t]
	\centering
	\includegraphics[width=\textwidth]{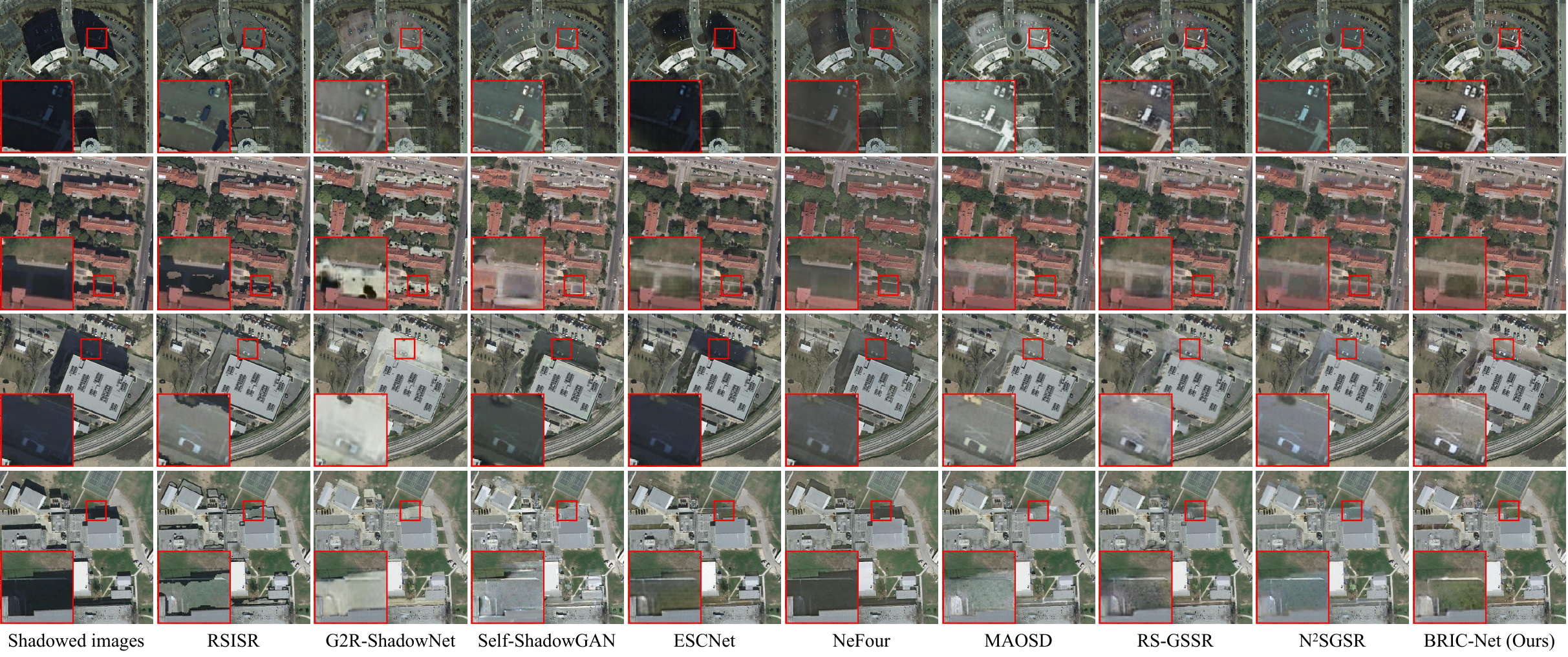}
	\caption{Additional cross-domain qualitative comparison on the
		real-world AISD dataset. The displayed cases contain gradual transitions, small
		objects, and narrow boundaries under domain shift. BRIC-Net reduces
		visible residual attenuation while maintaining plausible local
		texture and color continuity.}
	\label{fig:aisd}
\end{figure*}

Figures~\ref{fig:gta}, \ref{fig:syn}, \ref{fig:real},
and~\ref{fig:aisd} provide additional qualitative comparisons with
RSISR~\cite{Silva2018Near}, G2R-ShadowNet~\cite{Liu2021From}, Self-ShadowGAN~\cite{Jiang2023Learning}, ESCNet~\cite{Luo2023Evolutionary}, NeFour~\cite{Chi2024Neural}, MAOSD~\cite{Zhang2025Shadow}, RS-GSSR~\cite{Shao2025Generative}, and N$^2$SGSR~\cite{Bo2026SARU} on SRGTA, AeroDS-Syn, AeroDS-Real, and AISD,
respectively. The enlarged regions support closer examination of
residual attenuation, boundary continuity, local color consistency,
texture recovery, and structural preservation. Particular attention is
paid to dark residuals near shadow boundaries, bright or dark transition
bands, excessive correction of non-shadow regions, local color shifts,
over-smoothing, and the loss of small objects or narrow structures.

For the paired SRGTA and AeroDS-Syn datasets, the restored outputs can
be directly compared with the shadow-free ground truths, allowing the
fidelity of brightness, color, texture, and boundary transitions to be
examined. The real-world AeroDS-Real and AISD datasets do not provide
paired shadow-free references; their results are assessed in terms of
visual plausibility, local color consistency, structural preservation,
and cross-domain robustness. The following discussion focuses on
challenging cases from the four datasets, focusing on boundary
continuity, color consistency, texture fidelity, and structural
preservation.

In the first SRGTA example in Fig.~\ref{fig:gta}, the billboard casts multiple narrow shadows on the ground, including those from its thin supporting structures. Such slender shadow regions are challenging for methods that depend on fixed-scale morphological erosion and dilation. When the width of a shadow is comparable to or smaller than the structuring element, erosion may remove its valid interior, while dilation may merge it with regions or produce an overly broad correction support. These effects lead to incomplete restoration, discontinuous transitions, or excessive modification of neighboring non-shadow pixels. In BRIC-Net, the uncertain boundary band retains the transition support around these narrow shadows, reducing the risk that they are omitted. Consequently, BRIC-Net produces a smoother and more spatially continuous correction of the fine support shadows than the compared methods.

This smoothness is not obtained merely by suppressing high-frequency
content. In the second SRGTA example, the fine roof patterns remain
distinguishable after restoration and closely resemble those in the
ground truth. The recovery of gradual shadow transitions and
preservation of roof texture indicates that the method does not achieve
smoothness through indiscriminate over-smoothing. Instead, it maintains
structural fidelity while reducing shadow-boundary artifacts, supporting
the effectiveness of the proposed boundary-aware illumination--color
interaction.

Shadows cast over vegetation and exposed ground constitute another
challenge in remote sensing image deshadowing. These regions contain
strong local variations in reflectance, color, and texture, so a
spatially uniform or insufficiently constrained correction can introduce
artificial colors, amplified texture, or visible transition artifacts.
As illustrated by the AeroDS-Real examples in Fig.~\ref{fig:real},
several competing methods produce noticeable color discrepancies or
unnatural local responses after correcting shadows over vegetation and
soil. By contrast, BRIC-Net yields more coherent local colors and
smoother transitions between restored and surrounding illuminated
regions in the cases, while retaining recognizable surface structures.
Since shadow-free references are unavailable for AeroDS-Real, these
observations demonstrate improved visual plausibility rather than exact
recovery of the unknown surface radiometry.

The cross-domain AISD results in Fig.~\ref{fig:aisd} also expose a
remaining limitation of BRIC-Net. A visible color shift persists in the
first example, which is likely associated with differences in imaging
characteristics, scene appearance, and color distribution between the
AeroDS-Syn training data and the real AISD imagery. Nevertheless, the
method retains comparative advantages in structure-sensitive
regions. In particular, the grass textures in the second and fourth
examples remain more coherent after restoration, while the small
vehicles in the first and third examples preserve more distinct shapes
and local contrast. These properties are relevant to downstream remote
sensing tasks: preserving fine land-cover texture supports reliable
surface interpretation, whereas maintaining small-object boundaries and
contrast is important for object detection. Thus, although domain shift
continues to affect color recovery, the AISD results indicate that
BRIC-Net provides a favorable balance between shadow attenuation and
the preservation of task-relevant visual information.


\newpage

\newpage
\bibliography{deshadow}


\end{document}